\pdfoutput=1

\documentclass[11pt]{article}

\usepackage[]{EMNLP2022}

\usepackage{times}
\usepackage{latexsym}

\usepackage[T1]{fontenc}

\usepackage[utf8]{inputenc}

\usepackage{microtype}

\usepackage{inconsolata}

\usepackage{hyperref}
\usepackage{amsmath}
\usepackage{amsfonts}
\usepackage{graphicx}
\usepackage{subfigure}
\usepackage{breakurl}
\usepackage{multirow}
\usepackage{bm}

\usepackage{microtype}
\usepackage{booktabs}

\usepackage{xspace}
\newcommand\BASESIZE{$_{\small \textsc{base}}$\xspace}
\newcommand\LARGESIZE{$_{\small \textsc{large}}$\xspace}

\usepackage{algorithm} 
\usepackage{algpseudocode} 
\makeatletter
\renewcommand{\ALG@beginalgorithmic}{\small}
\makeatother
\algnewcommand\algorithmicinput{\textbf{Input:}}
\algnewcommand\algorithmicoutput{\textbf{Output:}}
\algnewcommand\INPUT{\item[\algorithmicinput]}
\algnewcommand\OUTPUT{\item[\algorithmicoutput]}

\newcommand\scalemath[2]{\scalebox{#1}{\mbox{\ensuremath{\displaystyle #2}}}}

\title{Improved Knowledge Distillation for Pre-trained Language Models via Knowledge Selection}

\author{Chenglong Wang\textsuperscript{1}, Yi Lu\textsuperscript{1}\footnotemark[1], \ Yongyu Mu\textsuperscript{1}, Yimin Hu\textsuperscript{1}, Tong Xiao\textsuperscript{1,2}\footnotemark[2] \ and Jingbo Zhu\textsuperscript{1,2} \\
	\textsuperscript{1}NLP Lab, School of Computer Science and Engineering, \\ Northeastern University, Shenyang, China\\
	\textsuperscript{2}NiuTrans Research, Shenyang, China\\
	\ttfamily{\{clwang1119,yilu001102\}@gmail.com}\\
	\ttfamily{\{xiaotong,zhujingbo\}@mail.neu.edu.cn}
}

\begin{document}
\maketitle

\begin{abstract}
Knowledge distillation addresses the problem of transferring knowledge from a teacher model to a student model.
In this process, we typically have multiple types of knowledge extracted from the teacher model.
The problem is to make full use of them to train the student model.
Our preliminary study shows that: (1) not all of the knowledge is necessary for learning a good student model, and (2) knowledge distillation can benefit from certain knowledge at different training steps.
In response to these, we propose an actor-critic approach to selecting appropriate knowledge to transfer during the process of knowledge distillation.
In addition, we offer a refinement of the training algorithm to ease the computational burden.
Experimental results on the GLUE datasets show that our method outperforms several strong knowledge distillation baselines significantly.
\end{abstract}
\footnotetext[1]{This work is done during the internship at Northeastern University NLP Lab.}
\footnotetext[2]{Corresponding author.}

\section{Introduction}
Pre-trained language models (PLMs) have significantly advanced state-of-the-art on various natural language processing tasks, such as sentiment analysis \cite{bataa2019investigation, baert2020arabizi}, text classification \cite{sun2019fine, arslan2021comparison}, and question answering \cite{yang2019end}.
Despite the remarkable results, PLMs have a large number of parameters which make them expensive for deployment \cite{yang2019end}.

To solve this problem, recent works resort to knowledge distillation (KD) \cite{hinton2015distilling} to compress and accelerate the PLMs \cite{sun2019patient, sanh2019distilbert, li2021dynamic, Liang2021mix}.
Its key idea is to transfer the knowledge from a large PLM (\textit{i.e.}, teacher model) into a lightweight model (\textit{i.e.}, student model) without a significant performance loss.
In this process, we typically have multiple types of knowledge extracted from the teacher model, such as response knowledge, feature knowledge, and relation knowledge \cite{gou2021knowledge}.
Recent works mainly focus on how the student model learns the transferred knowledge more efficiently, such as designing training schemes \cite{sun2019patient, mirzadeh2020improved, jafari2021annealing, li2021learning} or enriching task-specific data \cite{jiao2020tinybert, Liang2021mix}. 
However, few works consider how to make full use of the multiple types of knowledge into the training of the student model.

Our preliminary study (see Section \ref{sec-KeyforKD-exp1}) shows that not all of the knowledge is necessary for learning a good student model when conducting distillation with diverse knowledge.
Furthermore, inspired by dynamic KD \cite{li2021dynamic}, we assume that learning from certain knowledge at different training steps is beneficial to KD.
We conduct probing experiments to verify this assumption in Section \ref{sec-KeyforKD-exp2}.
Specifically, we repeat KD 200 times and randomly select knowledge at each training step.
We find that different distilled student models have a distinct performance. 
For example, the best student model is nearly 10\% higher than the worst one in terms of accuracy on the RTE dataset.
In addition, it notably exceeds the student model that learns from fixed knowledge. 
Based on our preliminary study, we have the following suggestion:
the distilled student model could achieve superior performance if it learns appropriate knowledge at each training step.
This suggestion motivates us to investigate how to select the appropriate knowledge for the student model during the process of KD.

Based on the above findings, we propose an actor-critic approach to selecting appropriate knowledge to transfer at different training steps \cite{bhatnagar2009natural}.
This approach first uses an actor-critic algorithm to implement a knowledge selection module via a long-term reward optimization.
This optimization can consider the influence of knowledge selection on future training steps.
Furthermore, we develop a multi-phase training approach that divides the distillation process into multiple phases and provides a particular reward at the phase end.
Compared to the usual actor-critic algorithm, it can ease the burden of computing rewards.
After that, we perform KD by employing the trained knowledge selection module to select knowledge at each training step.

We test the proposed approach on six GLUE datasets \cite{wang2018glue} which involve question answering, sentiment analysis, and textual entailment.
Experimental results show that our method significantly outperforms the vanilla KD method \cite{hinton2015distilling} and other competitive KD methods. 
Notably, results also show our BERT$_{6}$ (6-layer BERT) student model can achieve more than 98.5\% of the performance of teacher BERT\BASESIZE while keeping much fewer parameters ($\sim$61\%).
In addition, we prove that when armed with data augmentation, our approach can yield further improvements and be superior to TinyBERT \cite{jiao2020tinybert}, leading to a strong baseline with the data augmentation.

\section{Related Work}
Knowledge distillation (KD) \cite{hinton2015distilling} is widely used to compress and accelerate the pre-trained language models (PLMs) \cite{jiao2020tinybert, sun2020mobile, wang2020minilm, li2021dynamic}.
Its core idea is to transfer the knowledge from a large PLM (\textit{i.e.}, teacher model) into a lightweight model (\textit{i.e.}, student model). 
Recent works on KD could be classified into three groups.
The first group focused on utilizing various knowledge from the teacher model to distill the student model.
For example, multiple teacher models are leveraged to provide a more diverse and accurate knowledge for the student model \cite{liu2019multi, yuan2021reinforced}.
\citet{sun2019patient} exploited knowledge from intermediate layers of the teacher model during distillation.
Moreover, Weight Distillation \cite{lin2021weight} transfers the knowledge in parameters of the teacher model to the student model.
The second group tended to design effective strategies to facilitate the student model to learn the knowledge from different types of knowledge, such as enriching the task-specific data \cite{jiao2020tinybert, Liang2021mix} and the two-stage learning framework \cite{turc2019well, jiao2020tinybert}.
The third group that has attracted less attention generally explored appropriate data and teacher models for student models in the KD.
For example, \citet{li2021dynamic} proposed a data selection strategy that only selects some vital data for the student model according to its competency.
\citet{yuan2021reinforced} attempted to adjust the weights to the multiple teacher models during distillation.

Different from these methods, in this paper, we are concerned with how to select appropriate knowledge for student models during the process of KD.
To this end, we train an effective knowledge selection model via an actor-critic algorithm and employ it to select appropriate knowledge to transfer at each training step.


\section{Background}
\subsection{Knowledge Distillation}
In KD, the student model learns knowledge from the teacher model by mimicking corresponding behaviors.
In the distillation process, this mimicking can be implemented by minimizing the following loss function:
\setlength\abovedisplayskip{4pt}
\setlength\belowdisplayskip{6pt}
\begin{eqnarray}
	\scalemath{0.9}{
		\mathcal{L}_{KD} = \sum_{x \in \mathcal{X}}\mathcal{L}_{diff}(f^{S}(x), f^{T}(x))
	}
	\label{eq_kd}
\end{eqnarray}
where $\mathcal{X}$ is the training dataset, $x$ is an input sample, $f^{S}(\cdot)$ and $f^{T}(\cdot)$ are the functions of describing the behaviors of the student model and the teacher model, respectively.
$\mathcal{L}_{diff}(\cdot)$ is a loss function that evaluates the difference between their behaviors.
In the process of KD, due to many behaviors (\textit{e.g.}, extracting features) existing in the student and the teacher model, we typically have multiple types of knowledge.
Previous works mainly focus on designing elaborate $f^{S}(\cdot)$, $f^{T}(\cdot)$, and $\mathcal{L}_{diff}(\cdot)$ to encourage the student model to learn certain fixed knowledge better \cite{sun2019patient, jiao2020tinybert, Liang2021mix}.
Unlike them, in this work, we make full use of the multiple types of knowledge and select appropriate knowledge for the student model during the process of KD.

\subsection{Knowledge Types}
\label{sec:all-categories}
The student model can learn from various of knowledge extracted from the teacher model in KD.
According to \citet{gou2021knowledge}, the types of knowledge can often be \textit{Response Knowledge (ResK)}, \textit{Feature Knowledge (FeaK)}, and \textit{Relation Knowledge (RelK)}.
In addition, we consider a \textit{Finetune Knowledge (FinK)} derived from the training dataset on compressing PLMs.
The overview of these knowledge types is following:
\begin{itemize}
	\setlength{ \parskip}{-0.25em}
	\item \textbf{ResK}: ResK refers to the knowledge that the student model learns by mimicking the neural response of the last layer of the teacher model. It includes more information about the predicted results \cite{hinton2015distilling}.
	\item \textbf{FeaK}: FeaK denotes the feature representation that the student model learns from the outputs of the intermediate layers of the teacher model. Different from ResK, it contains more information about the calculation process of the teacher model.
	\item \textbf{RelK}: RelK represents the relationships between different layers or samples \cite{yim2017gift}. It focuses more on the relationship between model layers.
	\item \textbf{FinK}: FinK is the knowledge that the student model learns from ground-truth labels by finetuning on the training dataset \cite{hinton2015distilling, devlin2018bert}. Here, we also consider it as a knowledge type in KD.
\end{itemize}
\vspace{-0.3em}

In this work, we explore how to select appropriate knowledge from these types to transfer during the process of KD.
To ensure transferring the above types of knowledge to the student model, we design loss functions $\mathcal{L}_{ResK}$, $\mathcal{L}_{FeaK}$, $\mathcal{L}_{RelK}$, and $\mathcal{L}_{FinK}$, respectively.
The Appendix \ref{appendix:a} presents the design details of loss functions.

\section{Preliminary Study}
\label{sec-KeyforKD}
Due to its unique information and character, we assume that each type of knowledge has a different impact on KD results.
Based on this, we conduct preliminary studies to probe the relationship between knowledge and KD.

\begin{table}[t]
	\centering
	\scalebox{0.7}{
	\begin{tabular}{lcccc}
		\toprule[1.1pt]
		\multicolumn{1}{c}{\multirow{2}{*}{Knowledge}} & \multicolumn{2}{c}{RTE} & \multicolumn{2}{c}{SST-2} \\ \cline{2-5} 
		\multicolumn{1}{c}{}                           & BERT$_{6}$    & BERT$_{3}$    & BERT$_{6}$     & BERT$_{3}$     \\ \midrule
		FinK                               &65.8			&58.1 			&89.7			&88.0  \\ \hline
		ResK                               &\textbf{67.1}	&56.8   		&90.2			&87.5  \\ \hline
		FeaK                               &66.1			&57.8   		&90.4			&87.1  \\ \hline
		RelK                               &64.6			&\textbf{58.9}  &88.9			&88.2  \\ \hline
		FinK\&ResK                         &65.2			&55.6  			&90.0  			&\textbf{88.3}  \\ \hline
		FinK\&ResK\&FeaK                   &64.6			&56.7  			&90.0  			&87.0  \\ \hline
		FinK\&ResK\&FeaK\&RelK			   &64.5			&57.4  			&\textbf{90.7}  &87.7  \\
		\bottomrule[1.1pt]
	\end{tabular}
		}
	\vspace{-1mm}
	\caption{
		Accuracies (\%) of distilled student models on the RTE and SST-2 development sets.
		The teacher model is the BERT\BASESIZE.
		The student models are the 6-layer BERT$_{6}$ and the 3-layer BERT$_{3}$, respectively.
		Note that the knowledge is fixed during the process of KD.
	}
	\label{tab:fix-kd}
	\vspace{-4mm}
\end{table}

\begin{table}[t] 
	\centering
	\scalebox{0.85}{
	\begin{tabular}{clrrr}
		\toprule[1.1pt]
		\normalsize Task & \multicolumn{1}{c}{ \normalsize Method} & \multicolumn{1}{c}{\normalsize $S_{best}$} & \multicolumn{1}{c}{\normalsize $S_{worst}$} & \multicolumn{1}{c}{\normalsize $\Delta$} \\[-2pt] \midrule
		\multirow{3}{*}{RTE}   
			& Random-All 	& 68.4 & 58.3   & 10.1 \\[-2pt] \cline{2-5}
			& Random-One    & 67.2  & 58.9  & 8.7 \\[-2pt]  \cline{2-5}
			& Fixed-All 	&67.1 &64.5   & 2.6 \\[-2pt]  \hline
		\multirow{3}{*}{SST-2} 
			& Random-All  	& 91.3  & 86.1 & 5.2 \\[-2pt] \cline{2-5}
			& Random-One    & 90.7  & 87.9 & 2.8  \\[-2pt] \cline{2-5}
			& Fixed-All 	& 90.7 & 88.9   & 1.8 \\[-2pt] \bottomrule[1.1pt]
	\end{tabular}
	}
	\vspace{-2mm}
	\caption{
		Accuracies (\%) of $S_{best}$ and $S_{worst}$ on the development sets.
		Fixed-All denotes that we use the same knowledge to train the student model at each training step.
		Its results are taken from Table \ref{tab:fix-kd}.
	}
	\label{tab:random-kd}
	\vspace{-4mm}
\end{table}

\subsection{Different Knowledge for KD}
\label{sec-KeyforKD-exp1}
We conduct an experiment using different knowledge to distill the student models individually.
In the experiment, we tune the weights of each type of knowledge on the dev set, similar to tuning hyper-parameters.
Table \ref{tab:fix-kd} compares these distilled student models.
We can see that the student models have noticeable gaps in terms of accuracy while using different knowledge.
It indicates that \textit{not all knowledge is necessary for learning a good student model}.
Interestingly, we can also see that the appropriate knowledge changes as the student model or the dataset changes.
We conjecture that factors such as student capacity and sample complexity could affect transferring knowledge, which is consistent with \citet{stanton2021does}'s findings.

\subsection{Different Knowledge at Training Steps}
\label{sec-KeyforKD-exp2}
Inspired by dynamic KD \cite{li2021dynamic}, we assume that the appropriate knowledge may change dynamically because the student model and the sample keep changing during training.
We conduct probe experiments to verify this assumption.
We randomly select knowledge for the student model at each training step.
Here, we employ this random strategy for the training steps of all epochs (\textit{Random-All}) and the training steps of the first epoch (\textit{Random-One}), respectively.
To make experiments more general, both Random-All and Random-One are repeated 200 times.

We report the best and worst performance of the distilled student models denoted by $S_{best}$ and $S_{worst}$ in Table \ref{tab:random-kd}.
From the results of Random-All, we can see that there is a conspicuous gap in accuracy scores between $S_{best}$ and $S_{worst}$.
Notably, $S_{best}$ performs significantly better than $S_{worst}$ by 10\% on the RTE dataset.
We conjecture that $S_{best}$ means the student model is trained with the appropriate knowledge at most of the training steps, while $S_{worst}$ is the opposite.
In addition, we notice that $S_{best}$ achieves 68.4\% accuracy, which significantly surpasses all student models with a fixed knowledge.
It indicates that \textit{KD can benefit from certain knowledge at different training steps}.
This also motivates us to investigate the method for selecting appropriate knowledge.

For the results of Random-One, there is also a remarkable performance difference between $S_{best}$ and $S_{worst}$, though only the training steps of the first epoch in KD involves random selection of knowledge.
It shows that knowledge selection at the current training step affects the following training steps.
Therefore, we should consider the future influence while selecting appropriate knowledge.

\begin{figure*}[t!]
	\centering
	\subfigure[the architecture of the KSM]{
			\centering
			\includegraphics[scale=0.58]{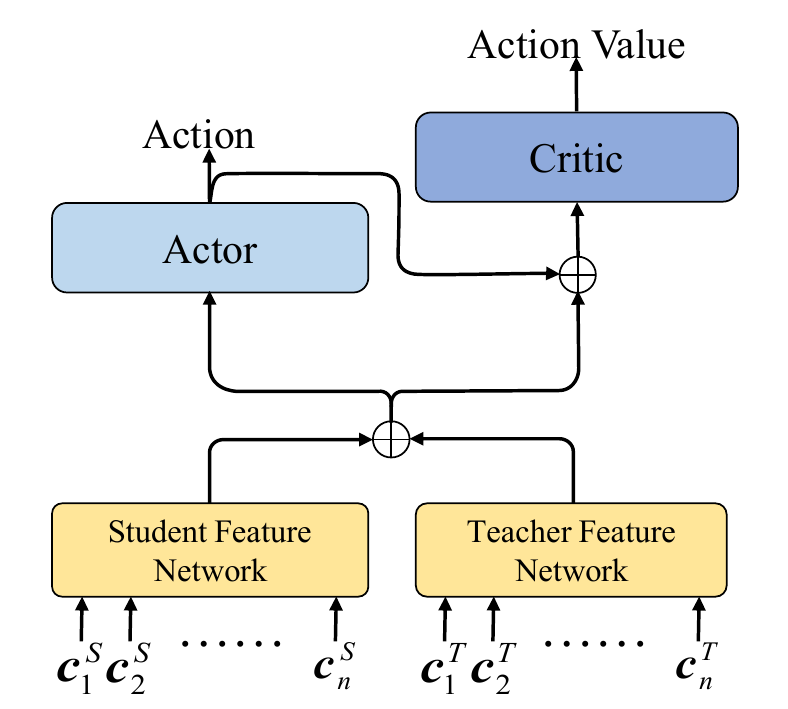}
			\label{sub-fig:actor-critic}
	}%
	\hspace{2mm}
	\subfigure[selecting appropriate knowledge for KD via the KSM]{
			\centering
			\includegraphics[scale=0.6]{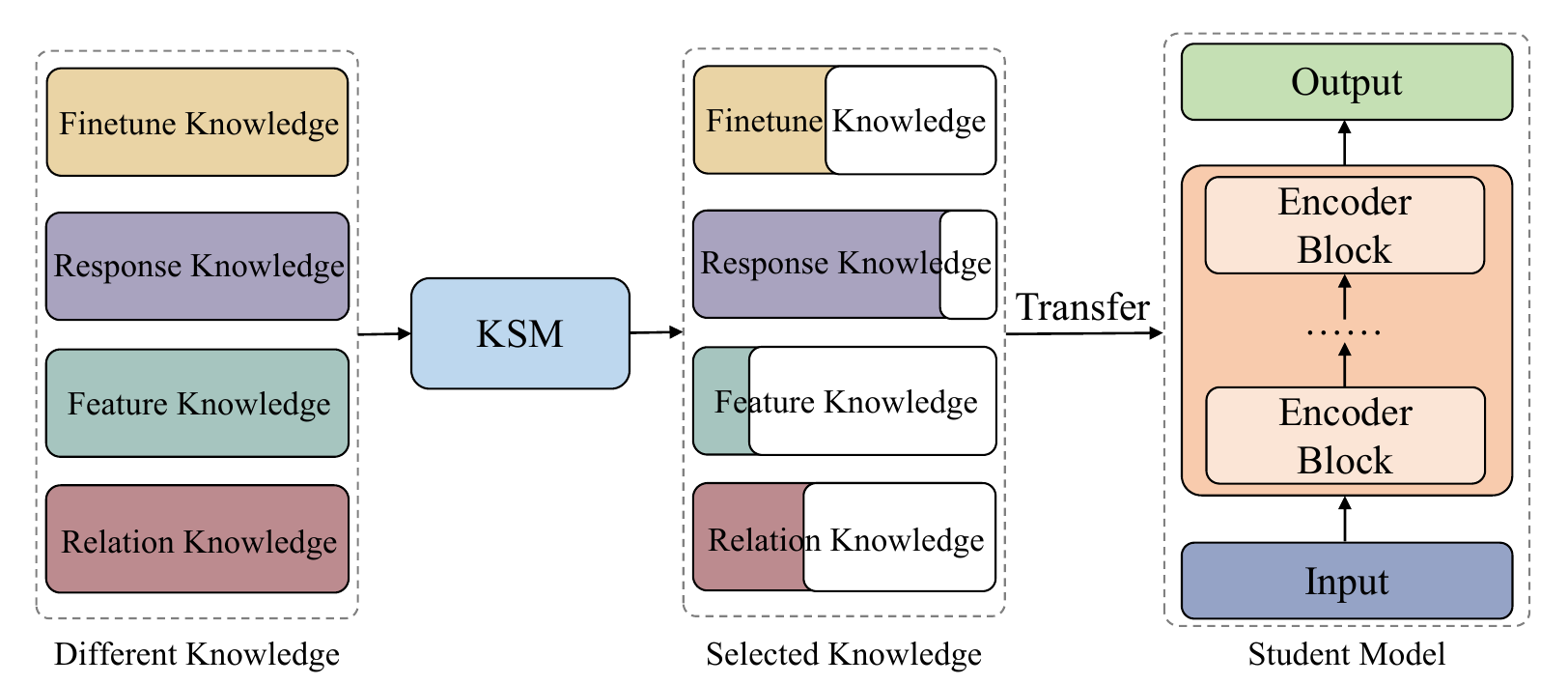}
			\label{sub-fig:main-figure}
	}%
	\vspace{-3mm}
	\caption{
	An overview of the proposed method.
	We use an actor-critic approach to design a knowledge selection module (KSM), which aims to select the appropriate knowledge to transfer during KD.
	Besides the soft action presented in Figure \ref{sub-fig:main-figure}, we also design the hard action for the actor module (see Section \ref{sec:ac-actor}). 
	}
	\label{fig:mainfig}
	\vspace{-5mm}
\end{figure*}

\vspace{-1mm}
\section{Method}
\label{sec:ackd}
\vspace{-1mm}
\subsection{Overview}
In this work, our goal is to select appropriate knowledge to transfer during the process of KD.
Considering the current selection may affect future training steps, we propose the actor-critic approach, which addresses the knowledge selection problem via a long-term reward optimization.
Figure \ref{fig:mainfig} gives an overview of our method.
Our method involves two stages: training a knowledge selection module (KSM) and distilling the student model with the trained KSM.
In the first stage, we use an actor-critic algorithm consisting of an actor and a critic to implement KSM.
The actor outputs an action that selects knowledge based on a given state.
The critic network predicts the action value, \textit{i.e.}, the sum of future rewards.
In the second stage, we employ the trained KSM to select appropriate knowledge and transfer them to the student model.

\subsection{Definitions}
In this work, we use an actor-critic algorithm to implement the KSM.
Here, we first describe some key concepts of the actor-critic algorithm, including state, action, and reward.

\vspace{-1mm}
\paragraph{State.}
At training step $t$, we use $s_{t}$ to denote the corresponding state.
Here, $s_{t}$ should comprise sufficient information for selecting appropriate knowledge.
To this end, we use the informative [CLS] embeddings of the last layers in the student and the teacher models \cite{devlin2018bert} to achieve it.
Specifically, for the $i$-th sample at training step $t$, we use $\bm{c}^{S}_{i}$ and $\bm{c}^{T}_{i}$ to denote the [CLS] embedding of the last layer of the student and the teacher model, respectively.
Then we utilize two trainable feature networks, $\mathcal{N}^{S}_{fea}$ and $\mathcal{N}^{T}_{fea}$, to extract the useful feature vector $\bm{v}$: 
\setlength\abovedisplayskip{4pt}
\setlength\belowdisplayskip{6pt}
\begin{eqnarray}
	\scalemath{0.9}{
		\bm{v}^{S}_{i} = \mathcal{N}^{S}_{fea}(\bm{c}^{S}_{i}),\ \ \ \bm{v}^{T}_{i} = \mathcal{N}^{T}_{fea}(\bm{c}^{T}_{i})
	}
	\label{eq:state}
\end{eqnarray}
where both $\mathcal{N}^{S}_{fea}$ and $\mathcal{N}^{T}_{fea}$ consist of a 2-layer multi-layer perceptron (MLP) network.
We concatenate all extracted feature vectors in given batch as $s_{t}$.

\vspace{-1mm}
\paragraph{Action.}
Given the state $s_{t}$, an action $a_{t}$ can be created, which is used to select the appropriate knowledge to transfer at training step $t$.
Here, we present soft and hard actions to select appropriate knowledge subtly.
The soft action determines how much to learn from each knowledge type, while the hard action selects one or more knowledge types for training the student model.

\vspace{-2mm}
\paragraph{Reward.}
Here we define the immediate reward $r_{t}$ as the cross-entropy loss difference on a development set after the student model is trained with selected knowledge.
See Table \ref{tab:diff-reward} in Appendix for a comparison of different rewards.

\subsection{Knowledge Selection Module}
\label{sec:ackd-method}
To improve the performance of the distillation model, we need to select the appropriate knowledge for transferring at different training steps.
For this purpose, we implement a knowledge selection module via the actor-critic algorithm which consists of an actor and a critic.
Here, we utilize long-term rewards to optimize it to consider the influence of knowledge selection on future training steps.

\subsubsection{Actor and Critic}
\paragraph{Actor.}
\label{sec:ac-actor}
In this work, the actor network $\mu_{\theta}$ is composed of a three-layer MLP with four output neurons.
It takes a state $s_{t}$ and outputs the soft action with a Sigmoid function:
\setlength\abovedisplayskip{4pt}
\setlength\belowdisplayskip{6pt}
\begin{eqnarray}
	a_{t} = \text{Sigmoid}(\mu_{\theta}(s_{t}))
\end{eqnarray}
where $a_{t}=\{ a^{t}_{1}, a^{t}_{2}, a^{t}_{3}, a^{t}_{4} \} $ contains four real values belonging to $[0, 1]$.
We use $a^{t}_{m}$ in $a_{t}$ to denote the percentage of the selected $m$-th type of knowledge\footnote{Here, we treat \textit{FinK}, \textit{ResK}, \textit{FeaK}, and \textit{RelK} as $1$-th, $2$-th, $3$-th, and $4$-th type of knowledge.}.
At training step $t$, the student model's loss is calculated as follows:
\begin{eqnarray}
	\setlength\abovedisplayskip{4pt}
	\setlength\belowdisplayskip{6pt}
	\scalemath{0.9}{
		\begin{aligned}
			\mathcal{L}^{soft}_{t} = a^{t}_{1}\mathcal{L}_{FinK}&+a^{t}_{2}\mathcal{L}_{ResK}+a^{t}_{3}\mathcal{L}_{FeaK}\\
			\ \ \ \ \ &+a^{t}_{4}\mathcal{L}_{RelK}
		\end{aligned}
	}
	\label{eq:kd-soft}
\end{eqnarray}
In addition, to directly select one or more types of knowledge, we obtain the hard action via a conditional function $g(\cdot)$: 
\setlength\abovedisplayskip{4pt}
\setlength\belowdisplayskip{6pt}
\begin{eqnarray}
	\scalemath{0.9}{
		g(a^{t}_{m}) =
		\begin{cases}
			1   &a^{t}_{m} \geq \lambda\\
			0   &otherwise
		\end{cases}
	}
\end{eqnarray}
where $\lambda \in [0,1]$ is a threshold value.
The student model's loss with the hard action is as follows:
\setlength\abovedisplayskip{4pt}
\setlength\belowdisplayskip{6pt}
\begin{eqnarray}
	\scalemath{0.9}{
\begin{aligned}
	\mathcal{L}^{hard}_{t} &= g(a^{t}_{1})\mathcal{L}_{FinK}+g(a^{t}_{2})\mathcal{L}_{ResK}+\\
	& \ g(a^{t}_{3})\mathcal{L}_{FeaK}+g(a^{t}_{4})\mathcal{L}_{RelK}
\end{aligned}}
\label{eq:kd-hard}
\end{eqnarray}
Here if $a^{t}_{m} \geq \lambda$, $m$-th type of knowledge is appropriate for the student model; otherwise, it is not.

\paragraph{Critic.}
The critic network $Q_{\phi}$ is also composed of a three-layer MLP.
It computes the action value $Q_{\phi}(s_{t}, a_{t})$ of a given pair $(s_{t}, a_{t})$, where $s_{t}$ and $a_{t}$ are concatenated as the input.
The action value is an estimation of the sum of rewards earned after the actor takes the action $a_{t}$ at the state $s_{t}$.
Here, we define the action value as the long-term reward, \textit{i.e.}, the performance gain of the student model from the training step $t$ to the training end.

\subsubsection{Optimization}

\paragraph{Actor Optimization.}
Following the deep deterministic policy gradient algorithm \cite{silver2014deterministic}, we optimize the long-term reward by updating the actor parameters via the sampled policy gradient:
\begin{eqnarray}
	\setlength\abovedisplayskip{4pt}
	\setlength\belowdisplayskip{6pt}
	\scalemath{0.9}{
		\nabla_{\theta}J \approx  \frac{1}{N} \sum_{t=1}^{N} \nabla_{\theta} Q_{\phi}(s_{t}, a_{t})
	}
	\label{eq:actor-gradient}
\end{eqnarray}

\paragraph{Critic Optimization.}
At each training step, we can only gain an immediate reward $r_{t}$.
However, the critic provides a long-term reward rather than an immediate reward.
Therefore, we employ Temporal-Difference \cite{tesauro1991practical} to approximate the actual long-term reward with $r_{t}$:
\setlength\abovedisplayskip{4pt}
\setlength\belowdisplayskip{6pt}
\begin{eqnarray}
	\scalemath{0.9}{
		Q^{*}_{\phi}(s_{t}, a_{t}) = \gamma^{t} r_{t} + Q_{\phi}(s_{t+1}, a_{t+1})
	}
	\label{eq:q^}
\end{eqnarray}
where $\gamma$ is a discount factor \cite{sutton2018reinforcement}.
Then we can give the optimization objective of the critic network through the approximate long-term reward:
\begin{eqnarray}
	\setlength\abovedisplayskip{4pt}
	\setlength\belowdisplayskip{6pt}
	\scalemath{0.9}{
		\mathcal{L}_{Q} = \frac{1}{N} \sum_{t=1}^{N} MSE(Q_{\phi}(s_{t}, a_{t}), Q_{\phi}^{*}(s_{t}, a_{t}))
	}
	\label{eq:q-optimization}
\end{eqnarray}
where $MSE(\cdot)$ is the mean squared error loss function, and $N$ is the number of training steps in KD.

\vspace{-1mm}
\paragraph{Feature Networks Optimization.}
During training KSM, we update the student feature network with the actor network and the teacher feature network with the critic network, respectively, in order to reduce the instability of training KSM.

\vspace{-0.5mm}
\subsubsection{Multi-Phase Training}
\label{sec:phase-training}
Due to a large number of training steps in some datasets, computing the reward via the loss on the development set can yield considerable costs during training the KSM.
For example, training with 5 epochs on the QQP dataset needs to compute the reward about 50,000 times in an episode.
To ease the computational burden, we propose a multi-phase training approach to train the KSM.
Specifically, we divide the complete KD process into \textit{multiple phases} where each phase contains $k$ training steps.
After the student model is trained on a phase, we treat the cross-entropy loss difference on a development set as the phase reward $r^{p}$.
It expresses the sum of all rewards in a phase.
Here we can use a phase optimization objective $\mathcal{L}^{p}_{Q}$ to train the critic network instead of the $\mathcal{L}_{Q}$:
\begin{eqnarray}
\setlength\abovedisplayskip{4pt}
\setlength\belowdisplayskip{6pt}
	\scalemath{0.9}{
		\mathcal{L}^{p}_{Q} = \frac{1}{N_{p}}\sum_{j=1}^{N_{p}}MSE(r_{j}^{p}, \sum^{e(j)}_{t=b(j)}\hat{r}_{t})
	}
\label{eq:phase-traing}
\end{eqnarray}
where $N_{p}$ is the number of phases in KD, $b(j)$ and $e(j)$ are the beginning and end training steps in the $j$-th phase.
The $\hat{r}_{t}$ is an estimated reward computed by Eq. \ref{eq:q^}:
\begin{eqnarray}
\setlength\abovedisplayskip{4pt}
\setlength\belowdisplayskip{6pt}
\scalemath{0.9}{
	\hat{r}_{t} =  \frac{Q_{\phi}(a_{t},s_{t})-Q_{\phi}(a_{t+1},s_{t+1})}{\gamma^{t}}
}
\end{eqnarray}

With the help of dividing phases, we can effectively relieve the computational burden. 
Generally, the conventional training of the actor-critic algorithm needs to compute rewards $N$ times, while our multi-phase training only calculates rewards $\lceil N/k \rceil$ times.

\begin{algorithm}[t]
	\caption{Our Method}
	\begin{algorithmic}[1]
		\INPUT the well-trained teacher model $\mathcal{M}_{T}$; the initial student model $\mathcal{M}_{S}$; training dataset $\mathcal{X}$
		\OUTPUT the distilled $\mathcal{M}_{S}$;
		\State \%\%\% \textit{the first stage} 
		\State Train the KSM via Algorithm \ref{alg:training-rl}; 
		\State
		\State \%\%\% \textit{the second stage}
		\For{$t=1$ to $N$}
		\State compute a state $s_{t}$ via Eq. \ref{eq:state};
		\State use an action $a_{t}=\mu_\theta(s_{t})$ to select appropriate \Statex \ \ \ \ \ \ knowledge;
		\State train $\mathcal{M}_{S}$ via Eq. \ref{eq:kd-soft} or Eq. \ref{eq:kd-hard};
		\EndFor
		\State return $\mathcal{M}_{S}$
	\end{algorithmic}
	\label{alg:ackd}
\end{algorithm}

\vspace{-2mm}

\begin{algorithm}[t]
	\caption{Training of the KSM}
	\begin{algorithmic}[1]
		\INPUT the well-trained teacher model $\mathcal{M}_{T}$; the initial student model $\mathcal{M}_{S}$; training dataset $\mathcal{X}$; the initial KSM (an actor $\mu_{\theta}$, a critic $Q_{\phi}$, two feature networks $\mathcal{N}^{S}_{fea}$ and $\mathcal{N}^{T}_{fea}$)
		\OUTPUT the trained KSM
		\For{$episode=1$ to $K$}
		\State reset $\mathcal{M}_{S}$;
		\State divide the whole KD process into $N_{p}$ phases; \label{line2}
		\For{$j=1$ to $N_{p}$}
		\For{$t=b(j)$ to $e(j)$}
		\State compute a state $s_{t}$ via Eq. \ref{eq:state};  \label{line5}
		\State sample an action $a_{t} = \mu_{\theta}(s_{t})$;  
		\State utilize the actions in the previous phases to \Statex \ \ \ \ \ \ \ \ \ \ \ \ \ \ \ \ \ \ compute $r^{e}_{t}$ via Eq. \ref{eq:exploration2} or Eq. \ref{eq:exploration1};
		\State compute an action value $Q_\phi(s,a)$;  \label{line8}
		\State train $\mathcal{M}_{S}$ via Eq. \ref{eq:kd-soft} or Eq. \ref{eq:kd-hard}; \label{line9}
		\EndFor
		\State compute $r^{p}_{j}$ via the loss difference;
		\State train the KSM via Eq. \ref{eq:q-optimization} and Eq. \ref{eq:phase-traing};  \label{line12}
		\EndFor
		\EndFor
		\State return KSM
	\end{algorithmic}
	\label{alg:training-rl}
\end{algorithm}


\begin{table*}[ht]
	\centering
	\scalebox{0.8}{
	\begin{tabular}{l|cc|ccccccc}
		\toprule[1.1pt]
		Method    & Student &\#Params & \begin{tabular}[c]{@{}c@{}}RTE\\(2.5k)\end{tabular} & \begin{tabular}[c]{@{}c@{}}MRPC\\(3.7k)\end{tabular} & \begin{tabular}[c]{@{}c@{}}MNLI-m/mm\\(393k)\end{tabular} & \begin{tabular}[c]{@{}c@{}}SST-2\\(67k)\end{tabular} & \begin{tabular}[c]{@{}c@{}}QNLI\\(105k)\end{tabular} & \begin{tabular}[c]{@{}c@{}}QQP\\(364k)\end{tabular} & Avg. \\[-1.5pt] \midrule
		BERT\BASESIZE (Teacher) &- 			&109.0M	& 69.3 & 87.6/83.5 			& 84.1/83.1  & 94.3 & 90.5 & 71.0/89.2 & 84.9 \\[-1.5pt] \midrule
		PKD                     & BERT$_{6}$&67.0M&65.5 & 85.0/\textit{79.9} & 81.5/81.0  & 92.0          & 89.0 & \textit{70.7/88.9} & 82.5      \\[-1.2pt]
		Dynamic KD              & BERT$_{6}$&67.0M&-    & 86.5/-    			& 81.8/81.0  & -    & -             & -         &-      \\[-1.2pt]
		DistilBERT              & BERT$_{6}$&67.0M&58.4 & 86.9/-    			& \textit{82.6}/81.3  & \textit{92.5} & 88.9          & 70.1/-    &-      \\[-1.2pt]
		Finetune                  & BERT$_{6}$&67.0M&65.2 & 85.1/79.2  		& 81.1/79.8  & 91.7  &87.1  & 69.4/88.2  & 81.8     \\[-1.2pt]
		Vanilla KD 				& BERT$_{6}$&67.0M&65.1 & 85.5/79.8 			& 82.4/81.6  & 91.4   &86.9           &70.0/88.4  & 82.2      \\[-1.2pt]
		Random-Hard  			& BERT$_{6}$&67.0M&65.1 &85.2/79.0 			& 82.1/81.3  & 91.2          & 86.7          & 69.8/88.3  & 82.0     \\[-1.2pt]
		Random-Soft  			& BERT$_{6}$&67.0M&64.8 &85.0/79.2   		&82.2/81.3   & 91.5          & 86.5        & 69.7/88.1  & 81.9      \\[-1.2pt]
		Our method (Hard Action) 		& BERT$_{6}$&67.0M&\textit{66.6} & \textit{87.7}/\textbf{82.2} & \textit{82.6/81.8} & 92.1 & \textit{89.0} & 70.5/\textit{88.9} & \textit{83.3}     \\[-1.2pt]
		Our method (Soft Action) 		& BERT$_{6}$&67.0M&\textbf{66.8} & \textbf{87.9/82.2} & \textbf{83.1/82.1} & \textbf{92.6} & \textbf{89.3} & \textbf{71.1/89.1} & \textbf{83.6}     \\[-1.2pt]
		\midrule[1.1pt]
		PKD 					& BERT$_{3}$&45.7M&58.2 			&80.7/\textit{72.5} 	&76.7/76.3 & 87.5  &\textit{84.7} & \textit{68.1}/87.8  & 77.7 \\[-1.2pt]
		Random-Hard  			& BERT$_{3}$&45.7M&54.3 			&80.7/71.8 				&76.8/76.1   &87.4  &82.9  &66.4/86.4   & 76.5     \\[-1.2pt]
		Random-Soft  			& BERT$_{3}$&45.7M&55.7 			&78.9/69.7 				&77.1/76.3  	&87.7  &82.3  &66.4/86.5   & 76.5    \\[-1.2pt]
		Our method (Hard Action) 		& BERT$_{3}$&45.7M&\textit{58.4} 	&\textit{81.4}/71.5 	&\textit{77.5/76.8}   &\textbf{88.4}   &84.6  &\textit{68.1/88.0} &\textit{77.9}      \\[-1.2pt]
		Our method (Soft Action) 		& BERT$_{3}$&45.7M&\textbf{58.7}  &\textbf{81.9/72.7} 	&\textbf{78.0/77.3}  &\textit{88.3}  &\textbf{85.1}  &\textbf{68.7/88.2}  &\textbf{78.3}     \\[-1.2pt]
		\bottomrule[1.1pt]
	\end{tabular}
	}
	\vspace{-2mm}
	\caption{
		Results from the GLUE test server.
		The best and second-best results for each group of student models are in \textbf{bold} and in \textit{italics}, respectively.
		The numbers under each dataset indicate the corresponding number of the training dataset. 
		For MRPC and QQP, we report F1/Accuracy. 
		We also report the average accuracy for each dataset in the “Avg.” column.
		The results for Dynamic KD are achieved via the Uncertainty-Entropy strategy \cite{li2021dynamic}.
		The results of DistilBERT and PKD are taken from \citet{jiao2020tinybert} and \citet{sun2019patient}, respectively.
	}
	\label{tab:test-results}
	\vspace{-5mm}
\end{table*}

\subsubsection{Exploration Reward}
To speed up the training of the KSM, we also design an exploration reward $r^{e}_{t}$ at training step $t$, which encourages the actor to take more different actions \cite{tang2017exploration}.
For the soft action, we consider the similarity between actions as $r^{e}_{t}$:
\begin{eqnarray}
\setlength\abovedisplayskip{4pt}
\setlength\belowdisplayskip{6pt}
\scalemath{0.9}{r^{e}_{t} = \alpha\times(1-\frac{1}{k}\sum_{l=1}^{k}Sim(a_{t}, a_{l}))}
\label{eq:exploration2}
\end{eqnarray}
where $Sim(\cdot)$ is a cosine similarity function, and $a_{l}$ is the $l$-th action in the previous phase.
In addition, we calculate exploration reward $r^{e}_{t}$ for hard action based on its repetition:
\begin{eqnarray}
\setlength\abovedisplayskip{4pt}
\setlength\belowdisplayskip{6pt}
\scalemath{0.9}{r^{e}_{t} = \alpha\times(1-\frac{count(a_{t})}{k})}
\label{eq:exploration1}
\end{eqnarray}
where $count(a_{t})$ is the number of times the action $a_{t}$ was taken in the previous phase, and $\alpha$ is a scale factor.
We provide the critic network with an additional optimization objective $\mathcal{L}^{e}_{Q}$ via $r^{e}_{t}$ computed by Eq. \ref{eq:q-optimization}.

\vspace{-1.5mm}
\subsection{Knowledge Distillation}
Algorithm \ref{alg:ackd} presents our overall method.
In the first stage, we train a KSM with an actor-critic algorithm. 
As described in Algorithm \ref{alg:training-rl}, we first divide the complete KD process into $N_{p}$ phases in an episode (line \ref{line2}). 
At each training step in a phase, we compute a state, an action, and an exploration reward (lines \ref{line5}-\ref{line8}) and train the student model with selected knowledge (line \ref{line9}).
Then, we compute the phase reward when the phase ends.
Subsequently, we optimize KSM via the $\mathcal{L}^{e}_{Q}$ and $\mathcal{L}^{p}_{Q}$ loss objectives computed by the Eq. \ref{eq:q-optimization} and Eq. \ref{eq:phase-traing}, respectively (line \ref{line12}).
This process iterates on episodes until the performance of the student model converges.
In the second stage, we adopt the trained KSM to select appropriate knowledge that is used to distill the student model.

\vspace{-1mm}
\section{Experiments}
\subsection{Datasets and Settings}
\vspace{-1mm}
\paragraph{Datasets.}
Following \citet{sun2019patient}, we conduct experiments on six GLUE datasets \cite{wang2018glue}: MNLI \cite{williams2017broad}, QQP \cite{chen2018quora}, QNLI \cite{rajpurkar2016squad}, SST-2 \cite{socher2013recursive}, MRPC \cite{dolan2005automatically}, and RTE \cite{bentivogli2009fifth}.

\vspace{-2mm}
\paragraph{Settings.}
We use BERT\BASESIZE as the teacher model and the BERT$_{6}$ and BERT$_{3}$ as the student models.
The details of training settings are shown in Appendix \ref{appendix:b}.
All experiments are repeated three times, and we report the average results over three runs with different seeds.

\vspace{-1mm}
\subsection{Baselines}
We compare the proposed method with \textbf{finetune} \cite{devlin2018bert}, \textbf{vanilla KD} \cite{hinton2015distilling}, and other competitive pre-trained model KD methods such as \textbf{patient knowledge distillation (PKD)} \cite{sun2019patient}, \textbf{DistilBERT} \cite{sanh2019distilbert}, and \textbf{Dynamic KD} \cite{li2021dynamic}.

In addition, we also design \textbf{Random-Soft} and \textbf{Random-Hard} baselines to evaluate the effectiveness of our method.
Random-Soft and Random-Hard denote that the KSM randomly takes a soft and hard action, respectively.

\subsection{Main Results}
We submit our model predictions to the official GLUE evaluation server, and the results are summarized in Table \ref{tab:test-results}.
First, compared with all baselines, our method can achieve optimal results while distilling the student models BERT$_{6}$ and BERT$_{3}$ on all the datasets.
Second, compared with the Random-Hard (or Random-soft), we can observe that our method achieves significant improvements in the F1 and the accuracy scores.
It demonstrates that the KSM system outperforms the baselines that randomly select knowledge to use.
Third, the BERT$_{6}$ student model distilled by our method with soft action significantly outperforms the popular vanilla KD by a margin of at least 1.4\%.
We attribute this to the fact that KD benefits from selected knowledge at different training steps.
Fourth, our method achieves the BERT$_{6}$ student model with $\sim$61\% parameters and achieves a similar performance compared with the teacher model BERT\BASESIZE.
For example, on the QQP dataset, the BERT$_{6}$ student model trained by our method with soft action has 89.1\% accuracy, which only is 0.1\% away from the teacher model.
In addition, compared with the teacher model BERT\BASESIZE, the BERT$_{3}$ student model has only 41\% parameters while maintaining 92\% performance.

Furthermore, we further investigate the performance gain on two different actions. 
From the results, we can find that the soft action performs better than the hard action, though both can contribute to our method to achieve an improvement over baselines on most of the datasets.
One potential explanation might be that the soft action space is larger than the hard action space.
Therefore, the soft action is more likely to explore a better knowledge selection than the hard action.

\begin{figure}[t!]
	\centering
	\includegraphics[scale=0.42]{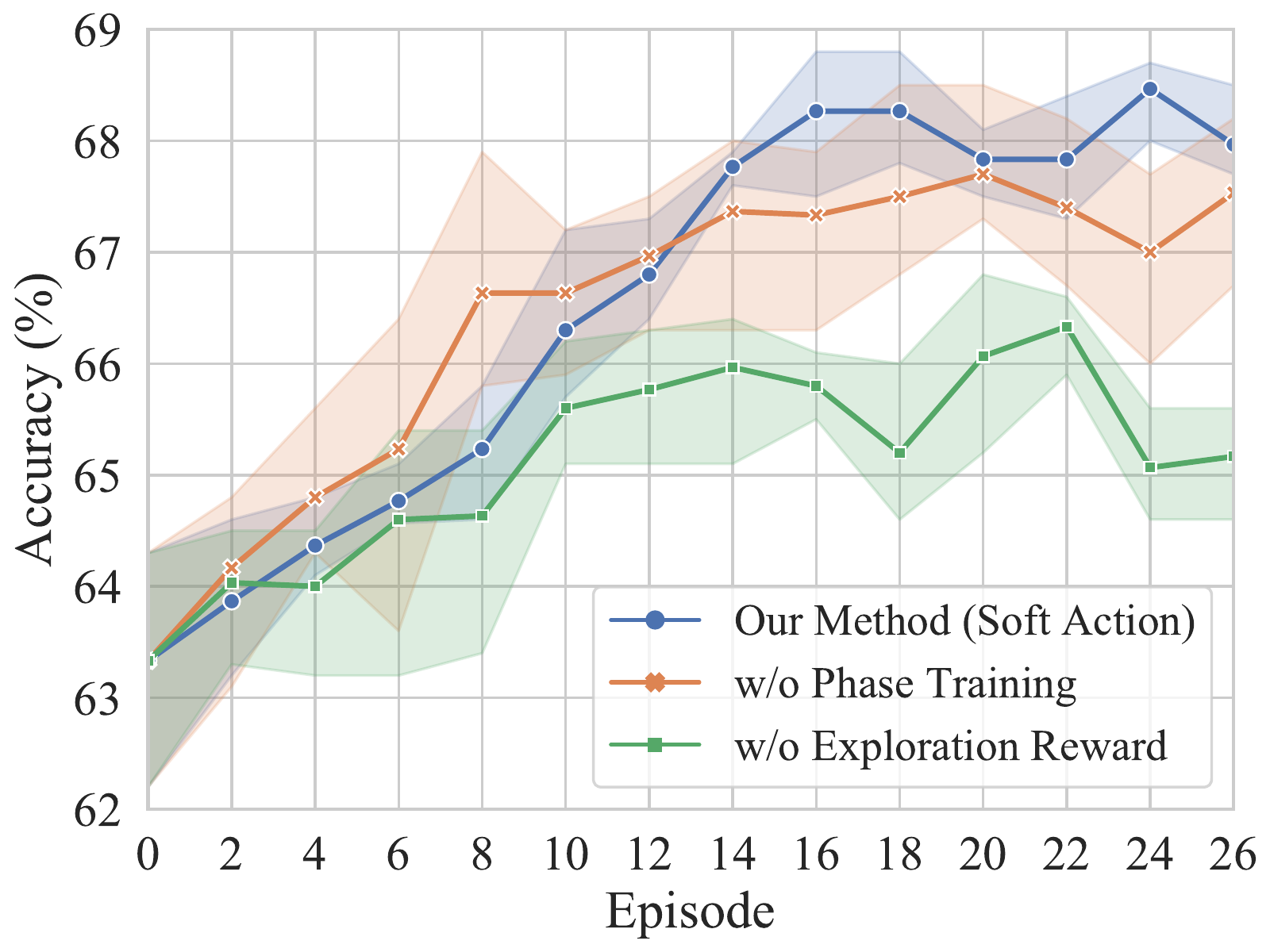}
	\vspace{-2mm}
	\caption{
		 Ablation study. 
		 We plot the mean accuracy of student models distilled with three different seeds on the RTE development set.
	 }
	\vspace{-1mm}
	\label{fig:ablation_rte}
 	\vspace{-4mm}
\end{figure}

\subsection{Ablation Study}
\label{sec:ablation-study}
We conduct an ablation study to explore the effects of the proposed multi-phase training and exploration reward on accuracy and efficiency. 
Figure \ref{fig:ablation_rte} shows the accuracy of student models distilled with the KSM on the RTE development set after removing multi-phase training or exploration reward.
We can see that using the multi-phase training can train a good KSM more likely, though the number of offering a reward is reduced.
We conjecture that the underlying reason is that the phase reward may be more accurate and stable than the immediate reward at each training step.
In addition, we can also observe that without the exploration reward, the KSM fails to explore the better knowledge selection more quickly.

\begin{figure}[!t]
	\centering
	\includegraphics[scale=0.42]{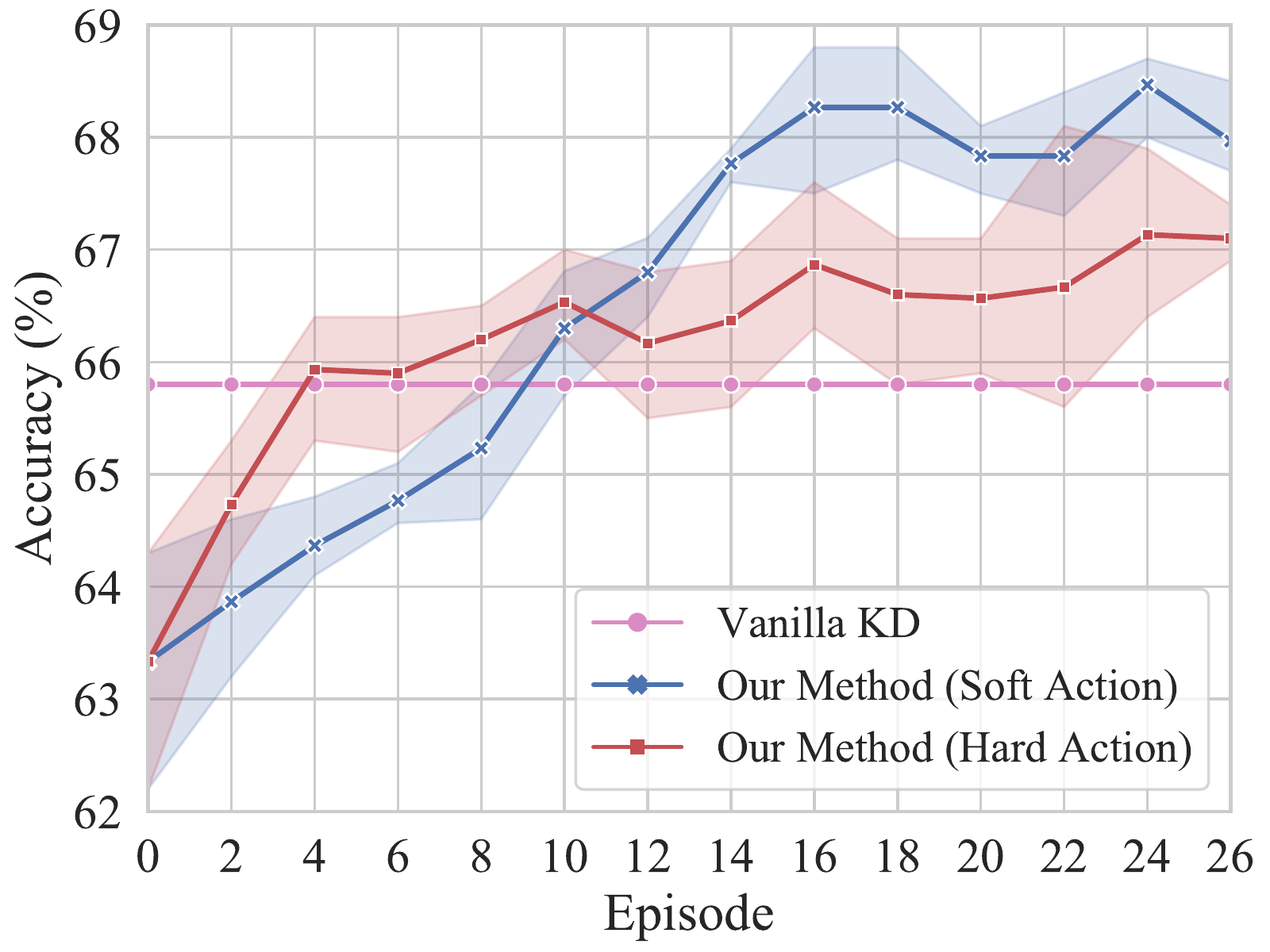}
	\vspace{-1mm}
	\caption{
		Comparison of the soft and hard actions.
		Our method can achieve faster KSM learning with the hard action and gain better results with the soft action.
	}
	\label{fig:soft-hard}
\end{figure}
\begin{table}[t] 
	\centering
	\scalebox{0.82}{
		\begin{tabular}{l|cccc}
			\toprule[1.1pt]
			Method      & \begin{tabular}[c]{@{}c@{}}RTE\\(52k)\end{tabular}  & \begin{tabular}[c]{@{}c@{}}MRPC\\(80k)\end{tabular} & \begin{tabular}[c]{@{}c@{}}SST-2\\(1,084k)\end{tabular} & \begin{tabular}[c]{@{}c@{}}QNLI\\(2,237k)\end{tabular}     \\[-1.2pt] \midrule
			Finetune      & 64.8 & 86.6/81.3  & 81.8  & 84.0             \\[-1.2pt]
			\hline
			Vanilla KD  & 67.3 & \textit{88.0/82.8}  & 91.2  & 87.0             \\[-1.2pt]
			TinyBERT    & \textit{70.0} & 87.3/82.6  & \textit{93.1}  & \textit{90.4}             \\[-1.2pt]
			\hline
			Our Method  & \textbf{70.4}     & \textbf{88.9/83.9} & \textbf{93.8} & \textbf{90.8}     \\[-1.2pt]
			\bottomrule[1.1pt]
		\end{tabular}
	}
	\caption{
		Results of our method with the augmented datasets from the GLUE test server.
		The results of TinyBERT are taken from \citet{jiao2020tinybert}.
	}
	\label{tab:augmented_dataset}
	\vspace{-4mm}
\end{table}
\vspace{-0.25mm}
\subsection{Discussion}
\paragraph{Performance on Data Augmentation}
Although the trained KSM can select the appropriate knowledge, the use of the fewer samples in the training dataset fail to provide the opportunity for the student model to learn them \cite{jiao2020tinybert, Liang2021mix}.
In such cases, we assume that if armed with the data augmentation, the student model could learn the knowledge assigned by KSM more sufficiently, thus achieving better performance.
To this end, we augment the training datasets (\textit{i.e.}, RTE, MRPC, SST-2, and QNLI datasets) about 20 times via the data augmentation procedure \cite{jiao2020tinybert}.
To make a fair comparison, we initialize our student model with the released BERT$_{6}$\footnote{\url{https://huggingface.co/huawei-noah/TinyBERT\_General\_6L\_768D}} distilled via General Distillation \cite{jiao2020tinybert}.
Table \ref{tab:augmented_dataset} compares our student model (soft action) with vanilla KD, finetune, and TinyBERT.
\begin{figure}[t]
	\centering
	\includegraphics[scale=0.42]{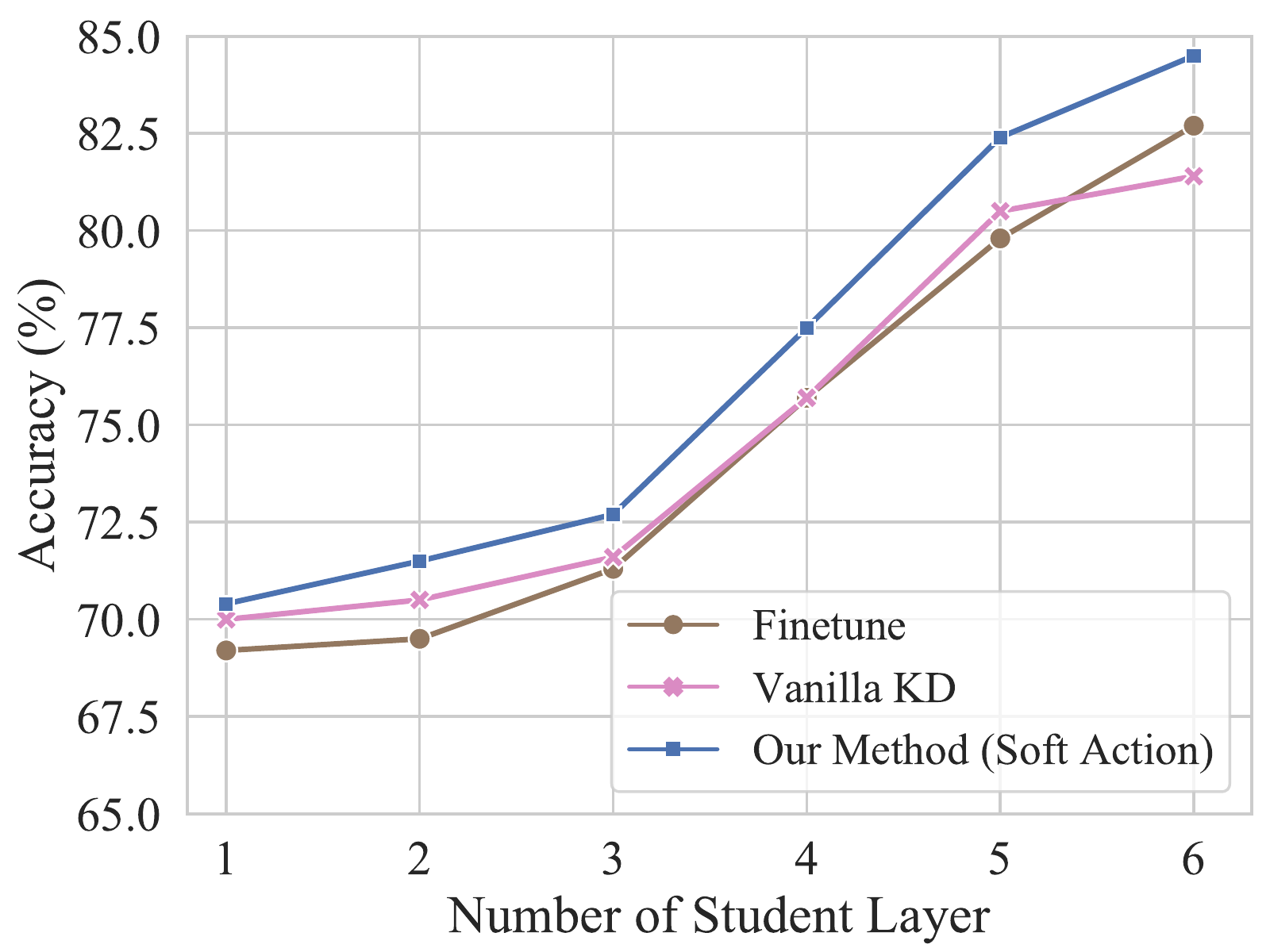}
	\vspace{-1mm}
	\caption{
		Performance of distilled student model with various layers on the MRPC development set.
	}
	\label{fig:diff-stu}
	\vspace{-2mm}
\end{figure}
The results show that our method is consistently better than all baselines.
It demonstrates that data augmentation can significantly improve the student models distilled by our method.
In addition, compared to the vanilla KD, our method can take advantage of the data more efficiently and transfer more knowledge to the student model.

This performance on data augmentation emphasizes that our method is orthogonal to the TinyBERT, as we address the knowledge selection problem during KD.
It also provides a suggestion that our method can be complementary to other KD methods, such as the MixKD \cite{Liang2021mix} and MiniLM \cite{wang2020minilm}.

\paragraph{Comparison of Soft Action and Hard Action}
\label{sec:comparison-soft-hard}
To intuitively present the effect of different actions on performance, we compare the soft and hard actions on the RTE dataset. 
Figure \ref{fig:soft-hard} shows the mean accuracy of the student models distilled with different action strategies on the development set.
It can be found that our approach can integrate seamlessly with different strategies.
Furthermore, we notice that the hard action can achieve faster KSM learning, while the soft action can achieve better results.
We can draw similar observations on the results of other datasets, \textit{e.g.}, on a relatively larger dataset QQP (see Figure \ref{fig:soft-hard-qqp} in Appendix).

\vspace{-1mm}
\paragraph{Performance on Different Student Models}
\label{sec:diff-stu}
To explore the performance of different student models, we distill student models with different layers and plot the performance to compare baselines (\textit{i.e.}, finetune and vanilla KD) in Figure \ref{fig:diff-stu}.
The results show that our method consistently outperforms the baselines while distilling the student model with various layers.
It indicates that our method can be adapted to the distillation of different student models well.

\paragraph{A Regularization Perspective on Knowledge Selection}
In practice, we consider that the knowledge selection can act as a regularization which prevents the co-adaptation \cite{grisogono2006co, sabiri2022mechanism} in KD, \textit{i.e.}, distilling a student model highly depends on a certain behavior of the teacher.
If the distilled student model receives the inappropriate knowledge from the dependent behavior of the teacher, it can significantly alter the performance of the student model, which is what might happen with overfitting \cite{hawkins2004problem, phaisangittisagul2016analysis}.
However, knowledge selection allows the student model to learn from multiple different behaviors properly at each training step during KD.
Thus, compared to the traditional KD methods that distill a student model via single or fixed knowledge, our method can ease the effect from the inappropriate knowledge from a behavior to prevent the co-adaptation. 

See more discussion in Appendix \ref{appendix:c}.


\vspace{-1mm}
\section{Conclusion}
\vspace{-1mm}
In this paper, we focus on making full use of the multiple types of knowledge into the distilling of the student model.
We have proposed an actor-critic approach to selecting appropriate knowledge to transfer during the process of knowledge distillation. 
To ease the burden of computing rewards during training, we propose a multi-phase training approach and an exploration reward.
Our experiments on GLUE datasets show that our method significantly outperforms several strong knowledge distillation baselines.

\section*{Limitations}
In this section, we discuss some limitations of this work as follows:
\begin{itemize}
	\item \textit{We train a model to select appropriate knowledge to transfer during the process of knowledge distillation}.
	In this process, training the model is time-consuming and resource-intensive.
	For example, on the RTE dataset, if we utilize one TITAN V GPU and set the distillation epoch to 5, it will take 40 minutes to train the model.
	\item \textit{It is difficult to scale the proposed approach to a large dataset}.
	The underlying reasons are:
	(\textit{i}) More training steps in the supersize dataset will make the knowledge selection question more difficult.
	(\textit{ii}) The larger dataset, the state is more complex, which may confuse the knowledge selection module.
\end{itemize}


\section*{Acknowledgements}
This work was supported in part by the National Science Foundation of China (Nos. 61876035 and 61732005), the China HTRD Center Project (No. 2020AAA0107904), Yunnan Provincial Major Science and Technology Special Plan Projects (Nos. 202002AD080001 and 202103AA080015), National Frontiers Science Center for Industrial Intelligence and Systems Optimization (Northeastern University, China. No. B16009) and the Fundamental Research Funds for the Central Universities.
We also thank anonymous reviewers for valuable feedback.

\bibliography{anthology,custom}
\bibliographystyle{acl_natbib}

\clearpage
\appendix
\section{Design Details of Learning Each Knowledge Type}
\label{appendix:a}
\subsection{Response Knowledge}
For learning the response knowledge, we use the vanilla KD loss function:
\begin{eqnarray}
	\setlength\abovedisplayskip{4pt}
	\setlength\belowdisplayskip{6pt}
	\scalemath{0.9}{ \mathcal{L}_{ResK} = \sum_{x \in B}KL(\sigma(\dfrac{f^{S}(x)}{\tau}), \sigma(\dfrac{f^{T}(x)}{\tau}))}
\end{eqnarray}
where $KL(\cdot)$ is the Kullback-Leibler divergence function, $\sigma(\cdot)$ is the softmax function and $\tau$ is a temperature hyper-parameter \cite{hinton2015distilling}.
For the sample $x$, $f^{S}(x)$ and $f^{T}(x)$ are the final outputs of the student model and the teacher model, respectively.

\subsection{Feature Knowledge}
For learning the feature knowledge, we utilize the PKD-Skip method \cite{sun2019patient}, where the student model learns the outputs of the assigned teacher model's layers.
The corresponding loss function can be defined by:
\begin{eqnarray}
	\setlength\abovedisplayskip{4pt}
	\setlength\belowdisplayskip{6pt}
	\scalemath{0.9}{ \scalemath{0.8}{ \mathcal{L}_{FeaK} = \sum_{x\in B} \sum_{i=1}^{L_{S}} \left\| \frac{\bm{h}^{S}_{x,i}}{\left\| \bm{h}^{S}_{x,i} \right\|_{2} } - \frac{\bm{h}^{T}_{x,I(i)}}{ \left\| \bm{h}^{T}_{x,I(i)} \right\|_{2} } \right\|_{2} } }
\end{eqnarray}
where $I(i)$ is the assigned indexes of the teacher layer for responding the $i$-th layer of the student model, $L_{S}$ is the number of layers of the student model, $\bm{h}^{S}_{x,i}$ and $\bm{h}^{T}_{x,i}$ are the [CLS] embeddings of $i$-th layer of the student model and the teacher model for the sample $x$, respectively.

\subsection{Relation Knowledge}
For learning the relation knowledge, we use the flow of solution procedure (FSP) method \cite{yim2017gift}.
Specifically, we first define the FSP matrix by:
\begin{eqnarray}
	\setlength\abovedisplayskip{4pt}
	\setlength\belowdisplayskip{6pt}	
	\scalemath{0.9}{  G(\bm{h}_{1}, \bm{h}_{2}) =  \frac{\bm{h}_{1} \times \bm{h}^\top_{2}}{\left| \bm{h}_{1}\right| } }
\end{eqnarray}
where $\bm{h}_{1}$ and $\bm{h}_{2}$ are the [CLS] embeddings.
Let $G^{S}(\cdot)$ and $G^{T}(\cdot)$ to denote the FSP matrices of the student model and the teacher model, respectively.
Based on this two FSP matrices, we can achieve the loss function for learning the relation knowledge:
\begin{eqnarray}
	\setlength\abovedisplayskip{4pt}
	\setlength\belowdisplayskip{6pt}
	\scalemath{0.9}{  
		\begin{aligned}
		 &\mathcal{L}_{RelK}=  \sum_{x \in B} \sum_{i=1}^{L_{S}-1} MSE(G^{S}(\bm{h}^{S}_{x,i}, \bm{h}^{S}_{x,i+1}), \\& \ \ \ \ \ \ \ \ \ \ \ \ \ \ \ G^{T}(\bm{h}^{T}_{x,I(i)}, \bm{h}^{T}_{x+1,I(i+1)})) 
	\end{aligned}
}
\end{eqnarray}
where $MSE(\cdot)$ is the mean squared error loss function.
Here we use the same layer selection strategy for the teacher model as the PKD-Skip method.

\subsection{Finetune Knowledge}
For learning the finetune knowledge, we use the loss function:
\begin{eqnarray}
	\setlength\abovedisplayskip{4pt}
	\setlength\belowdisplayskip{6pt}
	\scalemath{0.9}{ \mathcal{L}_{FinK}=-\sum_{x \in B}\mathbf{1}\{y=\hat{y}\}\log p(y|x) }
\end{eqnarray}
where $\mathbf{1}\{\cdot\}$ is the indicator function, $y$ is the predicted label and $\hat{y}$ is the ground-truth label of the sample $x$.
\section{Training Setups}
\label{appendix:b}

\subsection{Training the KSM}
We implement the KSM described in Section \ref{sec:ackd} based on PyTorch\footnote{\url{https://github.com/pytorch/pytorch}}.
Specifically, we use a 2-layer MLP to achieve the student and teacher feature networks.
The input layer size and output layer size are the [CLS] embedding size and 8, respectively.
Both the actor network and the critic network are composed of a 3-layer MLP.
The hidden size and output layer size in the actor are 256 and 4.
The hidden size and output layer size in the critic are 256 and 1.
In addition, there are six hyper-parameters for training the KSM.
The candidate values for these hyper-parameters are introduced in Table \ref{table:hy-trainingrl}.
During training the KSM, we can select the optimal hyper-parameter setups with the best development set accuracy. 
\begin{table}[ht]
	\centering
	\scalebox{0.88}{
		\begin{tabular}{l|c}
			\toprule[1.1pt]
			Hyper-parameter      & Value  \\[-1.5pt] \midrule
			Threshold value $\lambda$   & \{0.1, 0.2, 0.3\}  \\[-1.5pt] 
			Number of training steps $k$  & \{32, 64, 96, 128\} \\[-1.5pt]
			Scale factor $\alpha$    & \{0.1, 0.2\} \\[-1.5pt]
			Discount factor $\gamma$ & 0.98 \\[-1.5pt]
			Learning rate of the actor $\epsilon_{\phi}$ & 0.0002     \\[-1.5pt]
			Learning rate of the critic $\epsilon_{\theta}$ & 0.0002     \\[-1.5pt] 
			\bottomrule[1.1pt]
	\end{tabular}}
	\caption{Hyper-parameters for training the KSM.}
	\label{table:hy-trainingrl}
	\vspace{-4mm}
\end{table}

\subsection{Training the Teacher Model}
In this work, we denote the number of layers (i.e., Transformer blocks) as $L$, the hidden size as $H$, and the number of self-attention heads as $A$ for BERT.
We use the pre-trained language model \textbf{BERT\BASESIZE} ($L=12$, $H=768$, $A=12$, $Total Parameters=110M$) \cite{devlin2018bert} as the teacher model.
We initialize the BERT\BASESIZE with the \texttt{bert-base-uncased}\footnote{\url{https://github.com/google-research/bert}}.
The teacher models are trained with random seeds.
In addition, the sentence length is 128, and the learning rate is 5e-5.
We set the batch size and training epoch to 32 and 5, respectively.
Note that it is possible to plug-in any large pre-trained model such as BERT\LARGESIZE \cite{devlin2018bert} and RoBERTa \cite{liu2019roberta} in our method.

\subsection{Training the Student Model}
We primarily report results on two student models: \textbf{BERT$_{6}$} ($L=6$, $H=768$, $A=12$, $Total Parameters=67.0M$) and \textbf{BERT$_{3}$} ($L=3$, $H=768$, $A=12$, $Total Parameters=45.7M$).
We initialize the BERT$_{6}$ and BERT$_{3}$ with the bottom 6 and 3 transformer layers of BERT\BASESIZE, respectively.
In training the student model, we set the batch size to 32 and the sentence length to 128.
We select a well-trained student model through their scores among the learning rate set of \{2e-5, 3e-5, 5e-5\}.

\section{Discussion}
\label{appendix:c}
\begin{figure}[!t]
	\centering
	\includegraphics[scale=0.42]{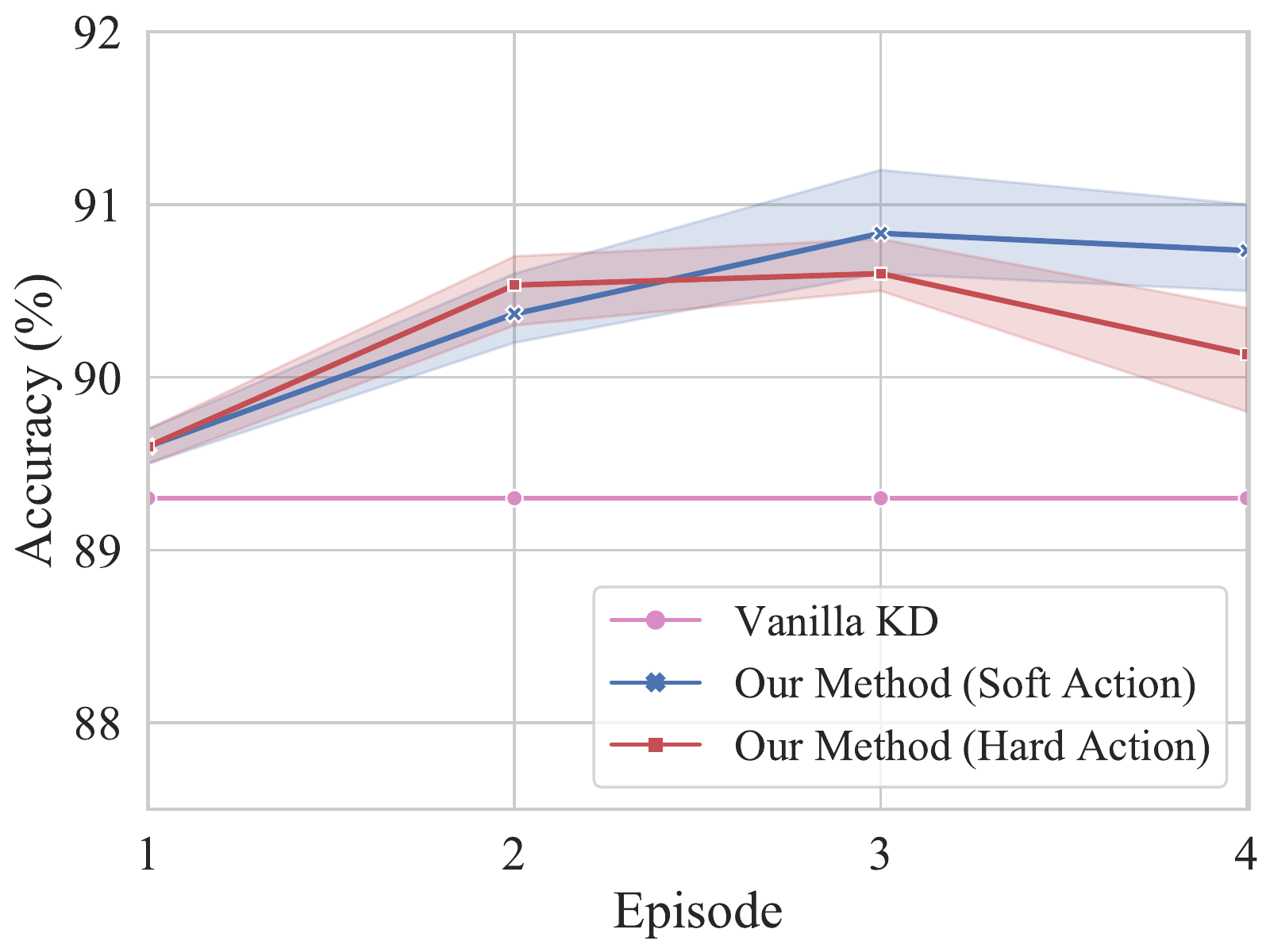}
	\vspace{-2mm}
	\caption{
		Comparison of the soft and hard actions on the QQP development set.
        For the QQP dataset, we can train a well-performance KSM only across two episodes.
        This is because a large dataset can provide more opportunities for updating the KSM in each episode. 
	}
	\label{fig:soft-hard-qqp}
\end{figure}
\begin{table}[t] 
	\centering
	\scalebox{0.9}{
	\begin{tabular}{l|cccc}
		\toprule[1.1pt]
		Reward      & RTE  & MRPC 		& QNLI & Avg.  	\\[-1.2pt] \midrule
		Loss    	& \textbf{68.4} & \textbf{89.3/84.7}  & \textbf{89.0} & \textbf{80.7}	\\[-1.2pt]
		F1  		& \textit{66.9} & 87.8/82.9  & \textit{88.1} & \textit{79.3}   \\[-1.2pt]
		Accuracy    & 66.7 & \textit{88.1/83.2}  & 87.3 & 79.1   \\[-1.2pt]
		\bottomrule[1.1pt]
	\end{tabular}
	}
	\caption{
	 	Results of different rewards for training KSM.
	}
	\label{tab:diff-reward}
	\vspace{-4mm}
\end{table}

\begin{figure}[!t]
	\centering
	\includegraphics[scale=0.42]{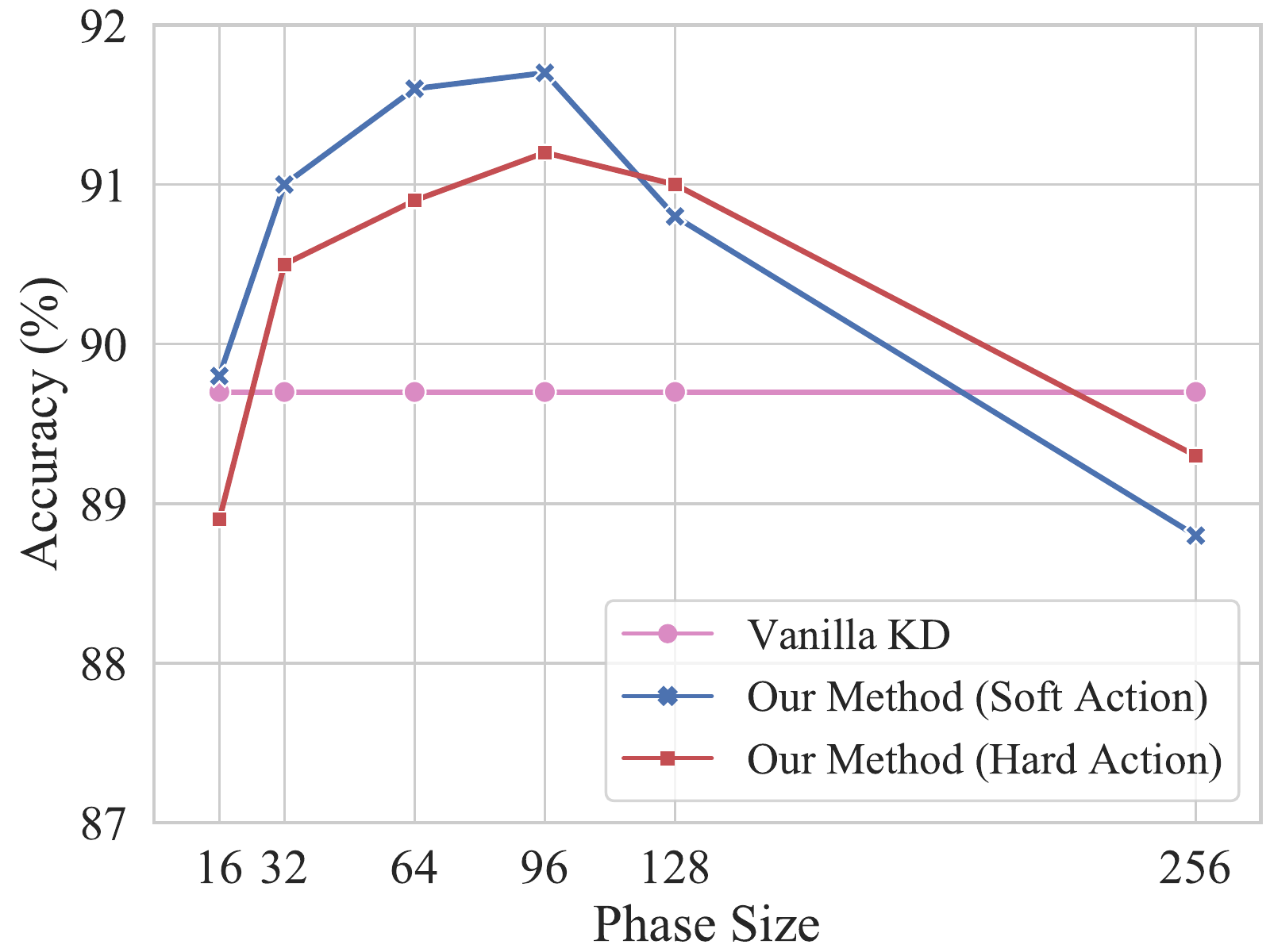}
	\vspace{-2mm}
	\caption{
		Performance of the student models distilled by our method with different phase sizes.
	}
	\label{fig:phase_size}
\end{figure}

\paragraph{Performance on Different Phase Sizes}
\label{sec:phase-size}
In Section \ref{sec:phase-training}, we divide the whole KD process into multiple phases. 
Then a question may arise about whether the phase size has an impact on the KSM performance.
To probe this question, we run our method to distill the student model BERT$_{6}$ with different phase sizes.
As shown in Figure \ref{fig:phase_size}, we observe that the excessive phase size can hurt the performance of the trained KSM, which we attribute to the deficiency of phase rewards.
In addition, the too small phase size is not beneficial to our method because its phase reward is similar to an immediate reward. 

\begin{figure}[t!]
	\centering
	\includegraphics[scale=0.42]{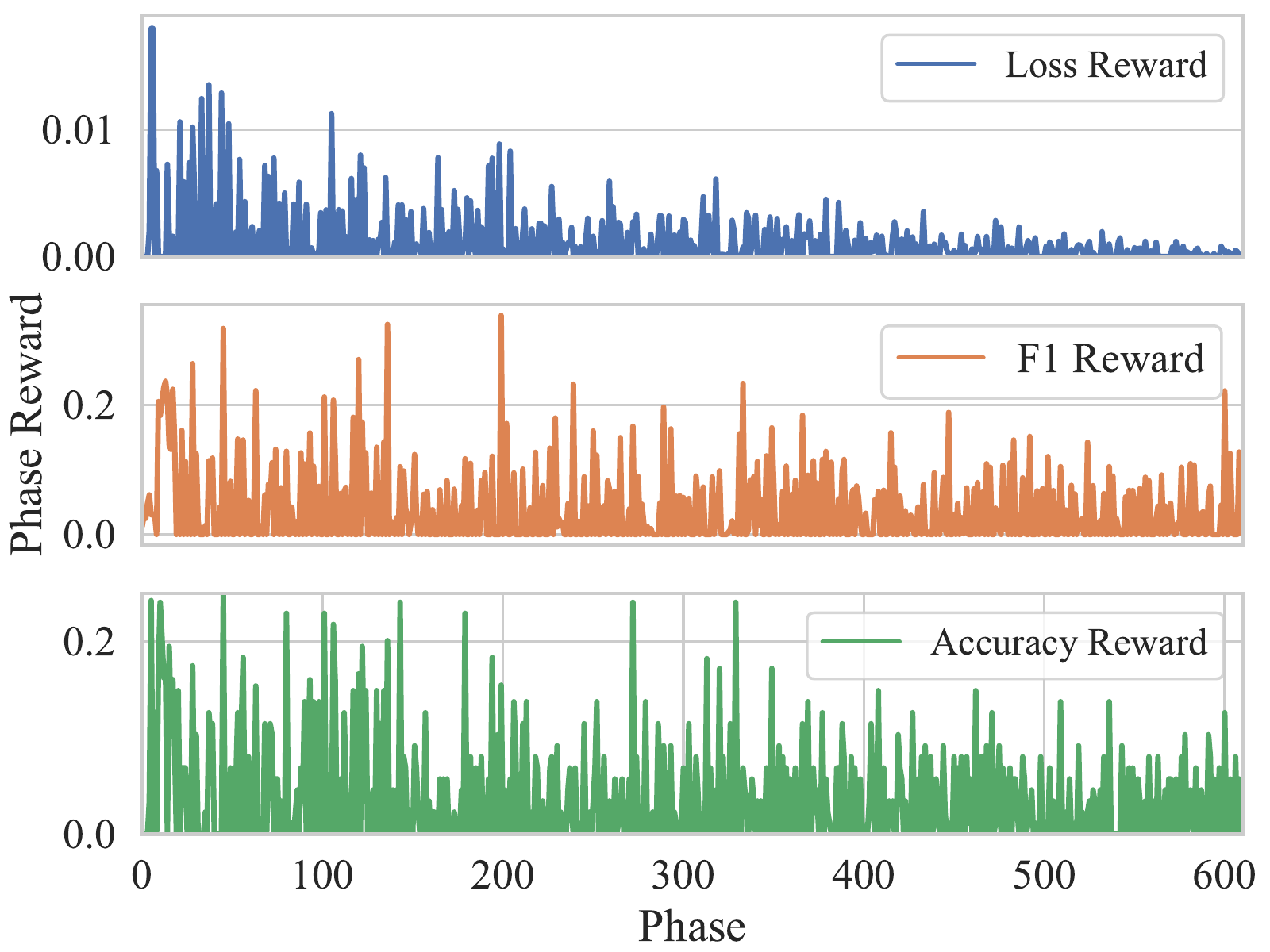}
	\vspace{-2mm}
	\caption{
		Phase reward comparison of the different rewards in one episode.
		The loss reward is more stable and smooth than the F1 reward and the Accuracy reward.
	 }
	\label{fig:diff-reward}
	\vspace{-4mm}
\end{figure}
\paragraph{Performance on Different Phase Rewards}
\label{sec:diff-reward}
We investigate the performance of our method while taking the task metrics (\textit{i.e.,} F1 or Accuracy) difference as the phase reward.
As shown in Table \ref{tab:diff-reward}, we observe that training KSM with a loss reward has the best performance overall.
To explore the reasons for this observation, we further compare the phase rewards of different rewards in one episode, as shown in Figure \ref{fig:diff-reward}.
The results show that the loss reward is more stable and smooth than the F1 and Accuracy reward.
Based on this, we conclude that the stability and smoothness give the loss reward the ability to train a better KSM.

\end{document}